\documentclass[twocolumn, switch]{article}
\usepackage{arxiv}

\usepackage[utf8]{inputenc}

\usepackage{cite}
\usepackage{amsmath,amssymb,amsfonts}
\usepackage{algorithmic}
\usepackage{graphicx}
\usepackage{textcomp}
\usepackage{xcolor}
\usepackage{subcaption}
\usepackage{multirow}
\usepackage{booktabs,threeparttable}
\usepackage{longtable}
\usepackage{array}
\usepackage{ragged2e}
\usepackage{arydshln}
\usepackage{authblk}

\usepackage{utf8}
\novocalize
\setfarsi
\settrans{farsi}
\yahnodots
\DeclareGraphicsExtensions{.pdf,.png}

\newsavebox{\spacebox}
\begin{lrbox}{\spacebox}
\verb*! !
\end{lrbox}
\newcolumntype{L}{>{\raggedright\arraybackslash}X}
\newcommand{\filliation}[5]{\affil[#1]{\textbf{#2}\vskip 0pt \textbf{#3}\vskip 0pt #4\vskip 0pt #5\vspace{10pt}}} 

\title{cycle text2face: cycle text-to-face gan via transformers}

\author[1]{Faezeh Gholamrezaie}
\author[2]{Mohammad Manthouri}

\filliation{1}{Dept. of Computer Science}{Shahed Univerisity}{Tehran, Iran}{faeze.gholamrezaie@shahed.ac.ir}
\filliation{2}{Dept. of Electrical and Electronic Engineering}{Shahed Univerisity}{Tehran, Iran}{mmanthouri@shahed.ac.ir}

\begin{document}

\twocolumn[ 
  \begin{@twocolumnfalse} 
  
\maketitle

\begin{abstract}

Text-to-face is a subset of text-to-image that require more complex architecture due to their more detailed production. In this paper, we present an encoder-decoder model called Cycle Text2Face. Cycle Text2Face is a new initiative in the encoder part, it uses a sentence transformer and GAN to generate the image described by the text. The Cycle is completed by reproducing the text of the face in the decoder part of the model. Evaluating the model using the CelebA dataset, leads to better results than previous GAN-based models. In measuring the quality of the generate face, in addition to satisfying the human audience, we obtain an FID score of 3.458. This model, with high-speed processing, provides quality face images in the short time. 

\end{abstract}
\vspace{0.35cm}
\end{@twocolumnfalse}]
    
\section{Introduction}

The text-to-image task is used in various applications such as face generation. The text-to-face model can be used in several ways, such as identifying criminals by eyewitnesses. Face generate has more details than other image production and requires more precision in manufacture. Since Photographic text-to-face synthesis is the combination of the two domains of Natural Language Processing (NLP) and Image Processing, the innovations of these two domains must be merged to create a suitable output. The goal of the mechanism  is to identify the features in the sentence and display them in the final face image.

In the first step of the text-to-face problem, vector representations of the text space must be provided. Text can be converted to a vector in two levels of word or sentence. In this process, the semantic similarity of the words must be preserved. Also, it pays attention to the role of words in sentence construction. To represent the vector space, algorithms such as Global Vectors for Word Representation (Glove) were presented, syntactic and semantic rules were considered, and a suitable structure was provided for embedding the text at the word level \cite{Pennington2014GloveGV}. The vector of a word that is expressed without attending its role in the sentence structure is incomplete.

Unsupervised learning models such as Skip-Thoughts and Quick-Thoughts have made significant progress, encoding input sentences in a fixed-dimensional display. In addition to considering the meaning of sentences, these sentence encoders provide outperform \cite{Kiros2015SkipThoughtV,Logeswaran2018AnEF}. The language representation models such as Bidirectional Encoder Representations from Transformers (BERT) and A Robustly Optimized BERT Pretraining Approach (Roberta) are designed to pre-train deep bidirectional, Which can provide a representation text \cite{Devlin2019BERTPO,Liu2019RoBERTaAR}. But the processing time of each of these models is long. To solve this problem, the sentence representation vector using Siamese BERT-Networks (Sentence-BERT), a change of the pre-trained BERT network, can deliver results in seconds \cite{Reimers2019SentenceBERTSE,Schroff2015FaceNetAU}. 

In the next step, it generates a realistic face by vector text that captures pertinent visual information. Recent advent and development of deep neural networks, and practical action was taken to provide quality images. Many models, including deep neural networks, such as StackGAN   \cite{Zhang2017StackGANTT}, StackGAN ++ \cite{Zhang2019StackGANRI}, AttnGAN \cite{Xu2018AttnGANFT}, MC-GAN \cite{Park2018MCGANMG}, MirrorGAN \cite{Qiao2019MirrorGANLT}, LeicaGAN \cite{Qiao2019LearnIA}, DM-GAN \cite{Zhu2019DMGANDM},  and MTC-GAN \cite{Zhang2021TextTI}, was introduced to provide good quality images. Each model tried to deliver images that were more relevant to the text. Newer models are more capable of creating higher quality and more accurate images than older models. None of the mentioned models produced faces and only created images of birds, flowers, or landscapes.

\begin{figure}
  \includegraphics[width=\linewidth]{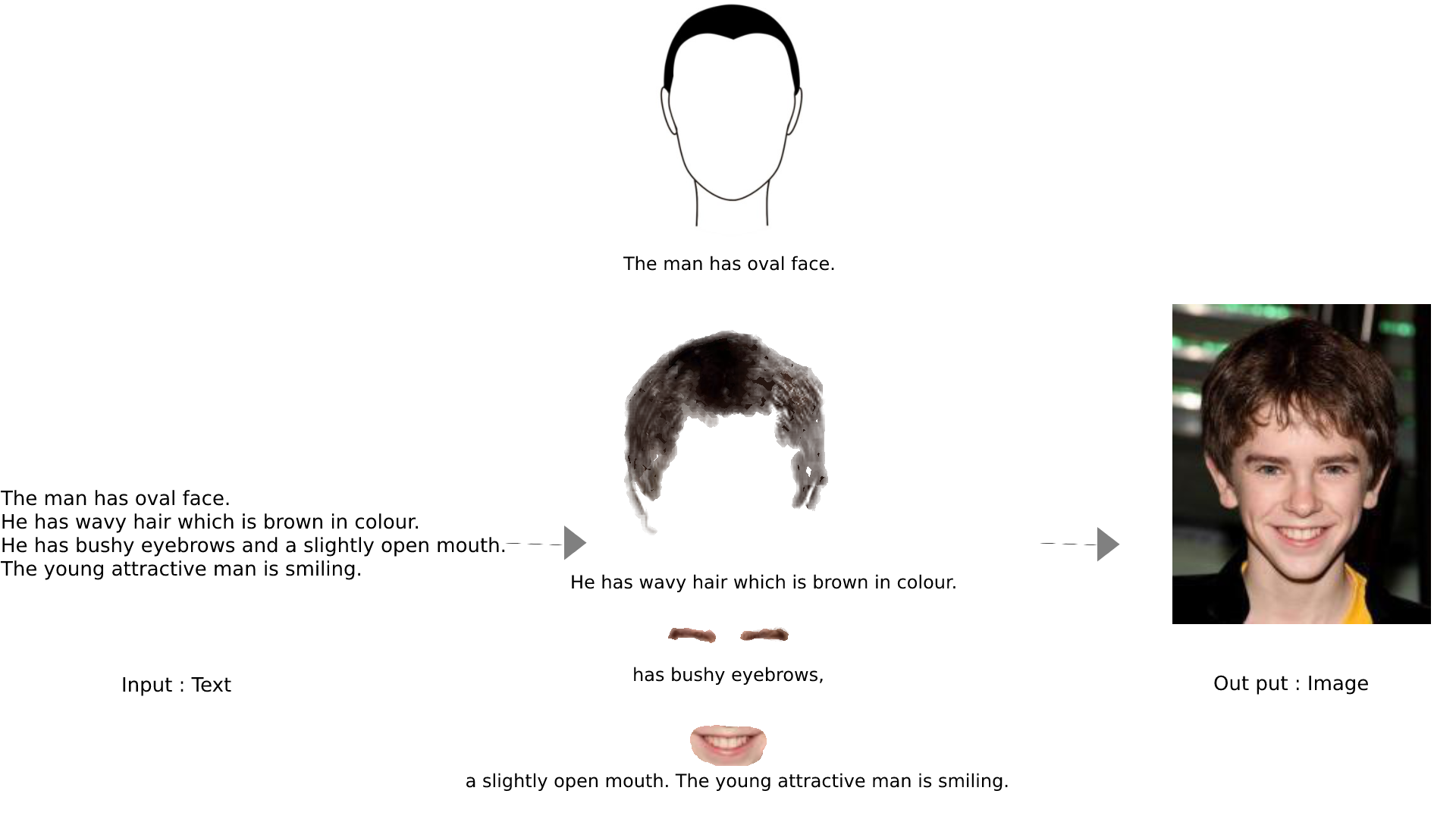}
  \caption{Each of the available subtitle sentences describes one of the components of the face. By drawing and juxtaposing different parts, the final image is equivalent to the desired face of the text.}
  \label{fig:img1}
\end{figure}

In the Text2Face model cycle, we use the large-scale CelebFaces Attributes(CelebA) dataset \cite{liu2015faceattributes}. But to produce a face, the relevant components of the face must be expressed in full detail in the qualified text, so we use text generate with \cite{Nasir2019Text2FaceGANFG} for the images (See Figure 1). We get meaningful embedding for each text, which maintains the similarities in the sentence. The sentence transformer is used to create the embedded text at the sentence level. To generate a face from the features mentioned in the sentence embedding, we use GAN.

The model introduced in this paper is a text-to-face cycle. This cycle includes an encryption-decryption model. In both the encoder and the decoder, an innovative innovation is used to turn text-to-face. In the encoder section of this model, we first obtain the representation vector of the caption using a transformer model, then convert it to a face image using GAN. In the decoder section, a caption describing the face is generated by extracting the facial features developed by the encoder step. Decoding in this model increases the accuracy of the model.

This article contains six sections. In the second part, the previous work done in the field of text-to-image is described. In the third section, definitions of the models used are provided.  In the fourth section, we introduce our decoder encoder model. In the fifth section, the oriented model is evaluated, and the results of this model are presented.

\section{Related Work}

There are various problems in the field of text-to-image conversion, which have led to different solutions. One of the problems in this field is the visualization of the story and the presentation of realistic images directly from a story text \cite{Zakraoui2021ImprovingTG}, In which the main role of the stories are human beings and the main challenge of human illustration is the descriptions expressed in the story. Methods compete with each other to improve results and provide quality images. These methods generate an image by extracting features from text well.
In 2009, displaying objects by their properties was the first step in this area \cite{Farhadi2009DescribingOB}.
To convert text to image, to create a Realistic image, you need to pay more attention to some features of the text \cite{Parikh2011RelativeA}.

The first method that was widely used in this field was zero-shot.In this method, by performing domain-related annotations, it achieved leading image recognition results \cite{Fu2014TransductiveME}. Considering the features and conditions in the text resulted in better quality output image \cite{Yan2016Attribute2ImageCI}. 

In 2016, new networks called Generative Adversarial Networks (GANs) were introduced. In the text-to-image problem, GANs, first advance the NLP and increased the accuracy of recognizing text features. Secondly, it revolutionized Image Processing and led to images that resemble reality \cite{Reed2016GenerativeAT}.

GANs is one of the deep neural networks, which was able to significantly improve the results of various problems raised in deep networks and machine learning, and artificial intelligence. The work done in the field of text-to-image conversion with the help of GANs can be divided as follows:

\subsection{text-to-image by text embedding in sentence-level}

In the first attempts to convert text to image, GANs provided 64x64 images, which were sharper than images produced by zero learning. In StackGAN networks, two steps were created to create an image that looked realistic and believable. In the first step, a low-resolution thumbnail was provided to provide an overview \cite{Zhang2017StackGANTT}. In the second stage, to create more clarity in the output image, more details were added to the output image of the first stage. Reinforcement learning was used to increase the accuracy of the images. Reinforcement learning increases the accuracy of model learning by providing an image of various directions and dimensions in the model training phase. Finally, 256 x 256 images were presented.

In the StackGAN ++ model, a tree structure containing multiple generators and detectors was considered \cite{Zhang2019StackGANRI}. Instead of producing a quality image of 64 * 64 or 256 * 256, they are made from different tree branches. Using this tree structure, and the two steps presented in the StackGAN model, was able to perform better in model training.

\subsection{text-to-image by text embedding in word-level}

In text-to-image, the text attribute must be specified. To extract the text features, one must pay attention to the words in the text \cite{Xu2018AttnGANFT}. AttnGAN networks do this task well. The AttnGAN network extracts individual objects and text attributes. After accurately identifying the things in the text, each object can be depicted in great detail, and the final image is produced with more parameters.

 A new step to increase the quality of results is to provide a text-related background in image generate. In addition to the details of objects and features, the MC-GAN network also provided a suitable environment for the image content \cite{Park2018MCGANMG}. 

The more attention yield to the word level, the closer the gap between the text and the image. In the MirrorGAN method, the new solution uses text-to-image as the reverse of image-to-text \cite{Qiao2019MirrorGANLT}. The generated image acts as a mirror and produces text similar to the original text. Using the mirror structure allows the reproduced images to be examined to show which of the text descriptions is correct, so this process causes the final output image to contain detailed text descriptions.

Each of the objects described in the text has a specific texture, color, structure, and arrangement. With increasing attention to words and using the LeicaGAN network, Taking these details into account will produce an image-realistic \cite{Qiao2019LearnIA}. image produced contains generalities, according to the words in the caption, and considering the structure of objects, a more accurate image is created.

In the stated methods text-to-image, first, a general initial image is created, then this image is corrected. Therefore, if the original image is inappropriate, the correction process can not provide an acceptable result. To solve this problem in the DM-GAN network, using dynamic memory in the GANs network and key-value memory, they could to use the memory to first store the information in the text in the memory structure . They set a dedicated key to retrieve it. In fact, by reading the features from memory, they corrected low-quality images. In this method, the words related to the image are selected dynamically using a memory writing gate \cite{Zhu2019DMGANDM}.

Pre-training is a multi-stage learning strategy that a simpler model is trained before the training of the desired complex model is performed. Pre-trained networks make it possible to provide accurate results in the shortest possible time. Bidirectional Encoder Representations from Transformers (BERT) is published by researchers at Google AI Language \cite{Devlin2019BERTPO}. BERT uses a Transformer, an attention mechanism that learns contextual relations between words in a text. In the text to image, appropriate text embedding can be obtained with the built-in BERT network, a model with this feature called Cycle GAN \cite{Tsue2020CycleTG}.

Creating simple images with less detail, such as a bird or flower, is easier than creating a face image. Face image production is complex due to the small and large details. Text-related face images were created using the MC-GAN network \cite{Nasir2019Text2FaceGANFG}. In the text data, text-to-face, various features of the face such as roundness of the face, small nose, eye color, etc. are propounded. In the MC-GAN network, 41 fixed features are provided for image generation. The number 1 is used for properties in the text and zero for properties that are not available. In this model, the details of the generated images are limited to the intended features. 

\begin{table}[h!]
\centering
\caption{Compare the results of different methods used to convert text to the image.}
\label{table:1}
\begin{tabular}{ |p{1.75cm}||p{1.25cm}|p{1cm}|p{3cm}|  }
 \hline
 \multicolumn{4}{|c|}{comparing results} \\
 \hline
 Networks & Quality & Images & Dataset \\
 \hline
 GAN & 64*64 &
\includegraphics[width=0.03\textwidth, height=8mm]{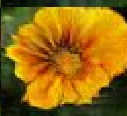}
\includegraphics[width=0.03\textwidth, height=8mm]{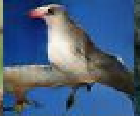}
\includegraphics[width=0.03\textwidth, height=8mm]{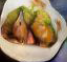}
\newline
 & Oxford-102Flowers \newline Caltech-UCSD  \newline MS COCO \newline \\
 StackGAN & 256*256 &
 \includegraphics[width=0.03\textwidth, height=8mm]{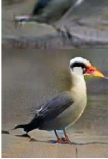}
\includegraphics[width=0.03\textwidth, height=8mm]{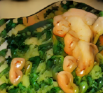}
\includegraphics[width=0.03\textwidth, height=8mm]{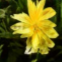}
\newline
 & Caltech-UCSD \newline MS COCO \newline  Oxford-102Flowers \newline\\
 StackGAN++ & 256*256 &  
 \includegraphics[width=0.03\textwidth, height=8mm]{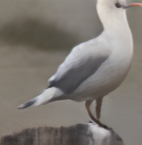}
\includegraphics[width=0.03\textwidth, height=8mm]{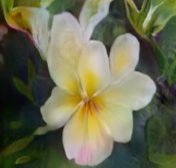}
\includegraphics[width=0.03\textwidth, height=8mm]{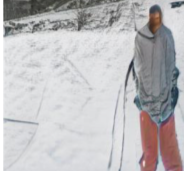}
\newline
 & Caltech-UCSD \newline Oxford-102Flowers  \newline MS COCO\newline\\
 MC-GAN & 128*128 & 
\includegraphics[width=0.03\textwidth, height=8mm]{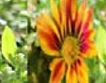}
\includegraphics[width=0.03\textwidth, height=8mm]{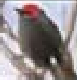}
\newline
 & Oxford-102Flowers \newline Caltech-UCSD \newline \\
 MC-GAN & 64*64 & 
\includegraphics[width=0.03\textwidth, height=8mm]{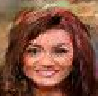}
\newline
 & CelebA\newline\\
 MirrorGAN & 128*128 & 
\includegraphics[width=0.03\textwidth, height=8mm]{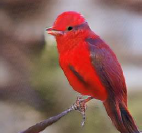}
\includegraphics[width=0.03\textwidth, height=8mm]{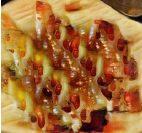}
\newline
 & Caltech-UCSD  \newline MS COCO\newline\\
 LeicaGAN & 256*256 &
\includegraphics[width=0.03\textwidth, height=8mm]{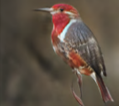}
\includegraphics[width=0.03\textwidth, height=8mm]{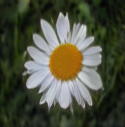}
\newline
 & Caltech-UCSD \newline Oxford-102Flowers \newline\\
 DM-GAN & 256*256 &
\includegraphics[width=0.03\textwidth, height=8mm]{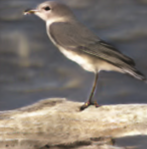}
\includegraphics[width=0.03\textwidth, height=8mm]{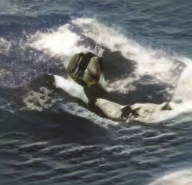}
\newline
 &  Caltech-UCSD  \newline MS COCO\newline\\
 Cycle GAN& 256*256 & 
\includegraphics[width=0.03\textwidth, height=8mm]{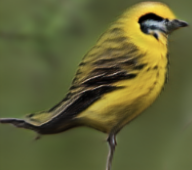}
\newline
 & Caltech-UCSD\newline\\
 \hline
\end{tabular}

\end{table}

Table \ref{table:1} Is compared the different methods text-to-image to the ones mentioned in terms of the type of models, dataset, quality, and output of the image.

\section{BACKGROUND}

This section describes the concepts used in the article. The first part provides a complete and comprehensive definition of GANs networks. The second subsection described the cycle GANs used in this paper. The use of pre-trained models improves the results. The third and fourth sections, respectively, pre-trained models and Sentence Transformers, the pre-training model used in this article, are discussed.

\subsection{Generative Adversarial Networks}
In 2014, a framework called Generative Adversarial Networks (GANs) was introduced \cite{Goodfellow2014GenerativeAN}. In this framework, two models under production model G and discriminator model D are teaching together. Generation Model G obtains the data distribution, and in return for this distribution, the discriminator Model D estimates the probability of its belonging to and similarity to the training data. The unique solution here is the G training method, which minimizes the G error probability while maximizing the probability of error D. In fact, this is a two-player game between G and D. The minimax calculation equation of this process is as follows:

\begin{equation}
\min _{G} \max _{D} V(D, G)=\mathbb{E}_{\boldsymbol{x} \sim p_{\text {data }}(\boldsymbol{x})}[\log D(\boldsymbol{x})]+\mathbb{E}_{\boldsymbol{z} \sim p_{\boldsymbol{z}}(\boldsymbol{z})}[\log (1-D(G(\boldsymbol{z})))]
\end{equation}

Which G is defined as follows:

\begin{equation}
G = \min\log (1-D(G(\boldsymbol{z})))
\end{equation}

In Equation (1), pg represents the distribution of the generator G on the x data. 
$p_{\boldsymbol{z}}(\boldsymbol{z})$
Displays the input noise variable. As a result, the G data space is equal to
$G\left(\boldsymbol{z} ; \theta_{g}\right)$
. Where G is a multilayer perceptron function with parameters
$\theta_{g}$
. D is also a multilayer perceptron function 
$D\left(\boldsymbol{x} ; \theta_{d}\right)$
with parameters $\theta_{d}$; the output D is a single scalar.

\subsection{Pre-trained Networks}

Although NLP covers a wide range of problems such as evaluating similarity, classifying, clustering, and summarizing texts, labeled and specific data on each of these problems are scarce \cite{Radford2018ImprovingLU}. The proposed solution is to use the old experience to improve the new models. Training does not have to start from scratch. In this method, the initialization of the model parameters is done by the results of the pre-trained model. Pre-training networks are deep neural networks that provide meaningful embedding of words.

\subsection{Sentence Transformers}

A general framework called Sentence Transformers was introduced to find sentence embedding whose primary purpose is to preserve sentence similarities so that the vector space of similar sentences is close together \cite{Reimers2019SentenceBERTSE}. Sentence Transformers include pre-trained models such as BERT / RoBERTa / XLM-RoBERTa and more. This framework tries to benefit from transfer learning with the help of pre-trained models.

\section{METHODOLOGY}
\subsection{Encoder}

The encoder part of Cycle Text2Face, firstly creating text embedding, and then generate a face image.
The embedding layer can be understood as a search table that maps from integers (for specific words) to dense vectors (embedded). We use sentence embedding to conceptualize the text. Embedding sentences allows us to use an efficient display, in which similar sentences have similar coding.  

Embedded dimension (or width) is a parameter you can use to test what works best for your problem, just as you would with a dense layer of neurons. When you create an embedding layer, the embedding weights are randomly initialized (like any other layer). During training, They are gradually adjusted through backward advertising. After training, the learned word embedding almost encodes the similarities between the words (as learned for the specific task your model taught).

We use the Sentence Transformers strategy and framework to embed sentences.
In this structure, the sentences that become numeric vectors are pretty meaningful, so that the appearance of numerical vectors can also be seen in the similarity between the sentences. Creates Siamese and triple networks in Sentence Transformers to coordinate between BERT / RoBERTa. The three Siamese networks are proportional in weight to the sentences produced (See Figure 2) \cite{Schroff2015FaceNetAU}.

\begin{figure}
  \includegraphics[width=\linewidth]{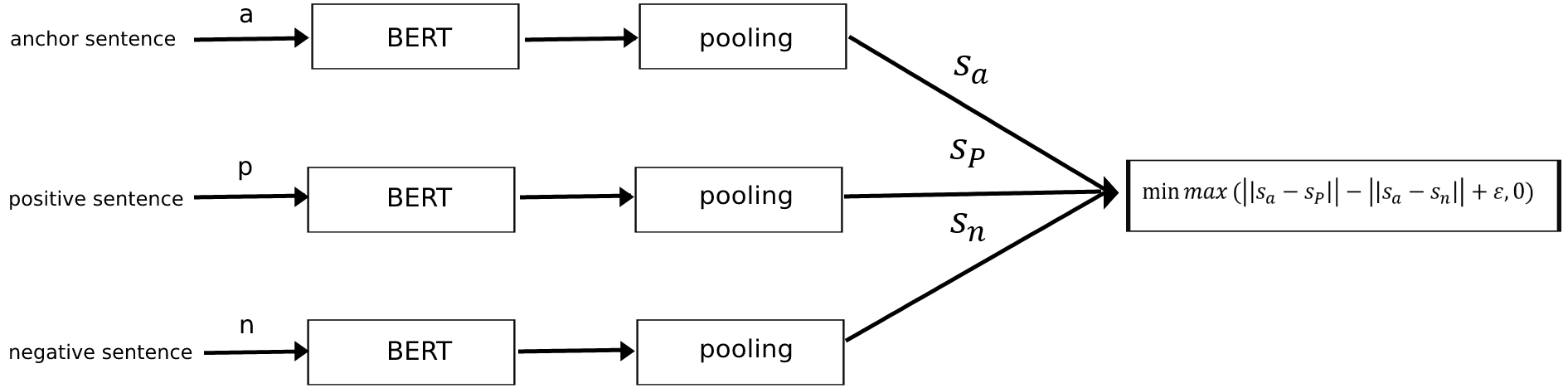}
  \caption{Sentence Transformers architecture. In this image, an anchor sentence is a, a positive sentence is p, and a negative sentence is n, An anchor sentence embedding $S_{a}$, a positive sentence embedding $S_{p}$, and a negative sentence embedding $S_{n}$. Finally, the loss function of this Siamese network is applied.}
  \label{fig:img2}
\end{figure}

The model Cycle Text2Face receives the subtitle of the face image as input and creates sentences embedded in the output. For each batch of input sentences, we create sentences that have a vector size of 512. This model adds noise with a dimension of 100 to vector text. Embedding Text Descriptions of face images are created so that similar semantic sentences have a similar embedding appearance.

\begin{figure*}
  \includegraphics[width=\textwidth,height=7cm]{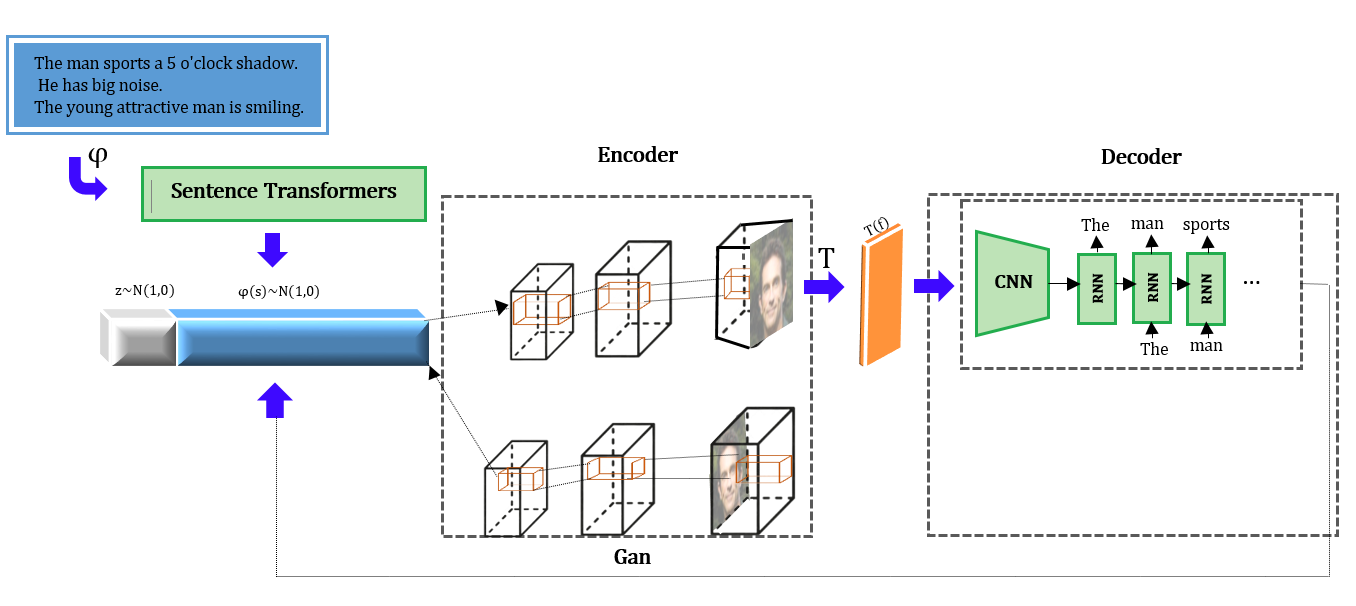}
  \caption{Cycle Text2Face Architecture.}
  \label{fig:img3}
\end{figure*}

We use a deep neural network, GAN, for text-to-face. In this model, we tasked GAN to generate face from text embedding Figure 3 shows the Cycle Text2Face architecture used in this article.

In the first step, the $\varphi$ function calculates the embedding of the face description text at the sentence level:

\begin{equation}
\mathrm{\bar{e}}=\varphi(\mathrm{s})
\end{equation}

where $\mathrm{s}$ is the sentence face descriptions and $\bar{e}$ is the sentence embedding. The $\varphi$ function receives the text describing the face, ie, $\mathrm{s}$, and by applying the Transformers model in question, generate embedding text $\mathrm{\bar{e}}$ with dimensions 512 that has a Gaussian distribution. 
Then, we concatenate the noise sequence with dimension 100 of embedded seq tensors. This text embedding structure has a Gaussian distribution (See Figure 4).

\begin{figure}[h]
    \centering
    \includegraphics[width=0.25\textwidth]{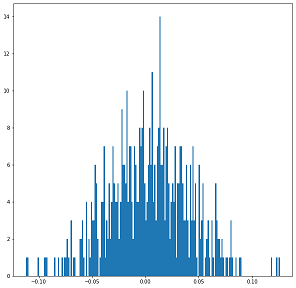}
    \caption{Embedding text in sentence level has a Gaussian distribution.}
    \label{fig:img4}
\end{figure}

GAN, the generator uses the 2D shifted convolution operator in this input matrix. Also, the activation function used by this generator network is Tanh. Fake-face generated by generator GAN calculate loss function criterion that measures the mean absolute error (MAE) between P-element in the fake-face and reality-face:

\begin{equation}
\mathrm{L_{  G_{e}}}=\frac{1}{P} \sum_{i=1}^{P}\left|r_{i}-\hat{f}_{i}\right|
\end{equation}

Where $L_{G_{e}}$ is the generator loss function of the GAN encoder model and, $P$ is the number of pixels in fake-face and reality-face images and, $\hat{f}_{i}$ is the i-th element in the fake-face, and ${r}_{i}$ is the i-th element in the reality-face.

After generating the fake-face by the generator, the discriminator detects the difference between the fake-face and the reality-face, updates the parameters accordingly. This discriminator network uses the LeakyReLU activation function. This discriminator uses the loss function criterion that measures the mean squared error (MSE):

\begin{equation}
\mathrm{L_{  D_{e}}}=\frac{1}{P} \sum_{i=1}^{P}\left(a_{i}-\hat{a}_{i}\right)^{2}
\end{equation}

Where $L_{D_{e}}$ is the discriminator loss function of the GAN encoder model and, $P$ is the number of pixels in fake-face and reality-face images and, ${a}_{i}$ is the i-th element in the activation-real and $\hat{a}_{i}$ is the i-th element in the activation-fake. Finally, network GAN leads to a face image. We turn the fake-face into a tensor by function T and send it to the decoder:

\begin{equation}
\mathrm{\bar{t}}=\mathrm{T}({ \hat{f} })
\end{equation}

$T$ is the function of converting the image to a tensor and, Where $\hat{f}$ is a fake-face image and, $\bar{t}$ is the tensor fake-face image. 

\subsection{Decoder}

The decoder extracts the image face features with a Convolutional Neural Network (CNN). We use the pre-trained model Inception-v3 to scale CNN networks. This model is a CNN architecture. Among the advantages of this model are improvements in Label Smoothing, Factorized 7 x 7 convolutions \cite{Szegedy2016RethinkingTI}.

Recurrent Neural Network (RNN) turns the output of the features CNN into descriptive words. We were using the CNN, and RNN network sequence on the fake-face tensor, for generated the fake-text. The decoder has a loss function Binary Cross-Entropy (BCE) for all N-words in the descriptor sentence:

\begin{equation}
L_{  G_{d}}=-\frac{1}{N} \sum_{i=1}^{N} \bar{t}_{i} \cdot \log \left(p\left(\bar{t}_{i}\right)\right)+\left(1-\bar{t}_{i}\right) \cdot \log \left(1-p\left(\bar{t}_{i}\right)\right)
\end{equation}

Where $L_{G_{d}}$ is the generator loss function of the GAN decoder model and, $N$ is the number of words in a sentence that describes a face and, $\bar{t}_{i}$ is the i-th element in the fake-face image tensor. The decoder loss function detects the difference between the fake-text and the reality-text and updates the parameters.

\section{EVALUATION AND RESULTS }

\begin{table}[h!]
\caption{Face images generated by Cycle Text2Face.}
\label{table:2}
\begin{center}
\begin{tabular}{  m{17em}  m{1.5cm} } 
 \centering
  & \begin{small} Face image \end{small} \\
\begin{small}
 The man sports a 5 o'clock shadow.
 
 He has big nose.
 
 The young attractive man is smiling.\end{small}& \includegraphics[width=0.095\textwidth,height=17mm]{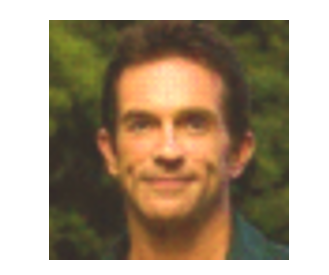} \\ 
\begin{small}
The woman has oval face and high cheekbones.

 She has wavy hair which is brown in colour.
 
 She has arched eyebrows.
 
 The smiling, young attractive woman has heavy makeup.
 
 She's wearing earrings, necklace and lipstick.\end{small}& \includegraphics[width=0.095\textwidth,height=17mm]{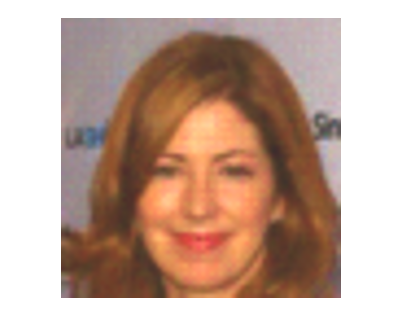} \\ 
\end{tabular}
\end{center}
\end{table}

Text-to-image problems provide new dimensions to the relationship between image processing and natural language processing. Text-to-face is presented as a subset of this domain. Text-to-face conversion problems have broad applications in enhancing human interactions by recognizing and producing faces through text.

The method introduced in this article, Cycle Text2Face, by applying the Cycle in its encoder-decoder model, can make face images from the text with precise details, natural color, and acceptable quality. Another advantage of the model we introduced is high-speed processing and delivering results in the shortest time.

The input of the Cycle Text2Face model is sentences that describe a face, in the form of a 512-dimensional matrix with a 100-dimensional noise matrix. Using a noise matrix prevents the model from overfilling. We used batches of the data set to train the model with a batch size of 64.

A continuous distribution is helpful in many applications such as process capability, control charts, and confidence intervals about point estimates. On occasion time to failure, data may exhibit behavior that a Gaussian distribution model well. For this reason, the embedded input sentences we have prepared have a Gaussian distribution of learning the best pattern.

 One of the best things about GANs is that they generate data that is similar to actual data. Because of this, they have many different uses in the real world. GANs are an exciting and rapidly changing field, delivering on the promise of generative models in their ability to generate realistic examples across a range of problem domains, most notably in text-to-image translation tasks such as translating the text of description face, and in generating realistic photo face  that even humans cannot tell are fake. In the encoder section, we use GANs to create an image of the face. Learning rate for generator in GANS was set to  0.0002 and for discriminator in GANs was 0.0001.We used Adam \cite{Kingma2015AdamAM} with $\beta_{1}$ = 0.9 and $\beta_{2}$ = 0.999 for both generator and discriminator.
 
 The advantages of CNN are local connection, weight sharing and, down-sampling. In the CNN model, each neuron is no longer connected to all neurons in the previous layer, but only to a small number of neurons. This reduces many parameters. Also, a set of connections can share the same weight, rather than having a different weight for each link, which reduces many parameters. Another feature CNN model, the Pooling layer, uses the principle of image local correlation to sub-sample the image, reducing the amount of data processing while retaining helpful information. Further, reduce the number of parameters by removing samples that are not important in the feature map. The main advantage of CNN compared to its predecessors is that it automatically detects the essential features without any human supervision. 
 
 In the decoder section, we first use the CNN model to identify the facial features in the image. Then we create the image descriptor sentences using the RNN model. RNN has a memory that remembers all information about what has been calculated. It uses the same parameters for each input as it performs the same task on all the inputs or hidden layers to produce the output. This reduces the complexity of parameters, unlike other neural networks.

Two of the results of model Cycle Text2Face are shown in Table 2. We evaluate the produced face images in two different sections. A first part is a machine, and the second is a human.

\subsection{Machine Evaluation}

There are currently several machine evaluation criteria for text-to-image production, the most prominent of Frechet Inception Distance (FID) \cite{Borji2019ProsAC}. FID is a measure of the distance between the Inception-v3 activation distribution for the generated and original image. The FID score is obtained from the following formula:

\begin{equation}
\mathrm{FID}(x, g)=\left\|\mu_{x}-\mu_{g}\right\|_{2}^{2}+\operatorname{Tr}\left(\Sigma_{x}+\Sigma_{g}-2\left(\Sigma_{x} \Sigma_{g}\right)^{\frac{1}{2}}\right)
\end{equation}

In this equation, x is the reality-face image, and g is the fake-face image. This equation model the data distributions for these features using a multivariate Gaussian distribution with mean $\mu$ and covariance $\Sigma$. Finally, it detects differences by distributing facial images.

Generate face images received an FID score of 3.458. Due to the more extended and more regular training, The Cycle Text2Face model obtained a good FID score. Although the FID criterion is the best machine criterion, but to show the superiority of our model over previous models, we obtained the machine evaluation criterion calculated in model Text2FaceGan. The evaluation criterion used in the Text2FaceGan model is IS that the Cycle Text2Face model was able to obtain a better number than Text2FaceGan in this evaluation (See Table 3).

\begin{table}[h!]
\caption{IS Comparison.}
\label{table:3}
\begin{center}
\begin{tabular}{  m{10em}  m{1cm} } 
\hline
\begin{center}Model\end{center}&\begin{center}FID\end{center}\\
\hline
Text2FaceGan \cite{Nasir2019Text2FaceGANFG} &1.4±0.7 \\
Cycle Text2Face & 1.20±0.08 \\
\hline
\end{tabular}
\end{center}
\end{table}

\subsection{Human Evaluation}

\begin{figure}[h]
    \centering
    \includegraphics[width=0.45\textwidth]{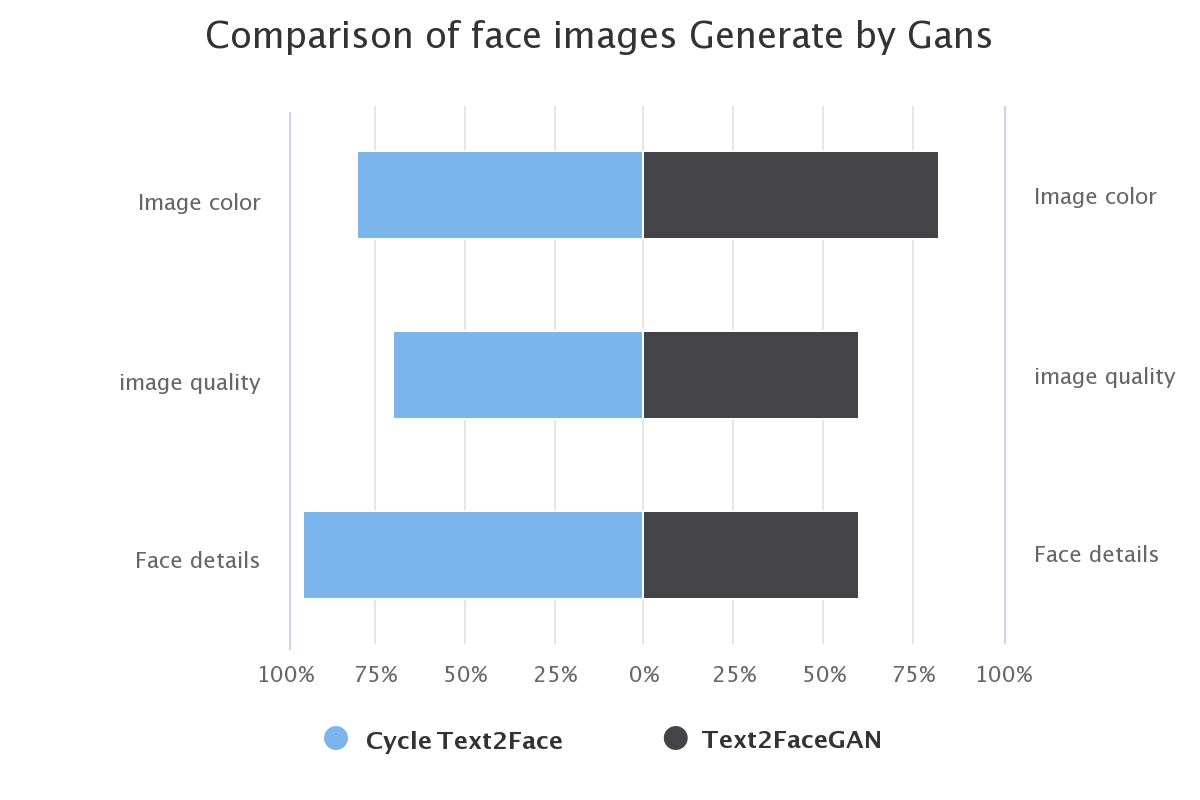}
    \caption{Human evaluation results. In this evaluation, three parameters of face detail, quality and color of the produced images were considered.}
    \label{fig:img5}
\end{figure}

In text-to-face evaluation, human evaluation is much more convincing. In this evaluation, we provided the human group with face images produced by Cycle Text2Face and model \cite{Nasir2019Text2FaceGANFG}.

The results of this evaluation are shown in Figure 5. As you can see in this image, the faces produced are evaluated with hree different parameters of face details, quality, and color. In this evaluation, each human audience gives each feature a score between 1 and 10. According to the results of this evaluation, model Cycle Text2Face has made good progress in producing details and has taken a positive step towards satisfaction in the other two features.

\section{CONCLUSION}
In this work, we presented an Encoder-Decoder model called Cycle Text2Face to accurately convert text to face image. To increase the accuracy of the model in receiving text, we used a sentence transformers model to create text embedding. The sentence transformer model is a pre-trained model that, by receiving the text, focusing on each sentence, provides the most appropriate text embedding according to the text of the sentence, so that similar sentences have the same text embedding. In the model Encoder section, we used GANs to convert text embedding to face image. In the Decoder section of the model, we convert the image of the generated face back to text.

The Cycle Text2Face model is a language-based model that creates the most meaningful embedding because it uses transformers with a sentence-level focus on the input text. The face images produced by this model have precise details due to the use of the GANs network because the Discriminator considers the smallest details of the face in comparison and reminds the Generator. Due to the use of cyclic structure and conversion of the generated image to text in this model, we were able to cover the gap between the two domains of face image and text. We considered the experiences of both NLP and Image processing domains using pre-trained models related to the same field so that we could better teach the model training phase.

To develop a text-to-face task, we suggest correcting the text of the subtitle description face, by predicting words that are misspelled or misplaced and convert them into face images. Also, we use annotations to produce higher-quality images. 
Model coverage can be to the extent that the text is understood in different languages. Face descriptions can be in any language; the new model can turn it into a face by identifying the desired language.
 Provide customer-based application systems based on text-to-image, such as story visualization systems, and Identify the face of the offender systems.

\bibliographystyle{unsrt}
\bibliography{references}
\vspace{12pt}

\end{document}